\documentclass[journal]{IEEEtai}

\usepackage{amsmath,amssymb}
\usepackage{graphicx}
\usepackage[colorlinks,urlcolor=blue,linkcolor=blue,citecolor=blue,breaklinks=true]{hyperref}
\usepackage{booktabs}
\usepackage{url}
\usepackage{xurl}
\usepackage{tikz}
\usetikzlibrary{shapes.geometric, arrows.meta, positioning, fit, backgrounds, calc}
\usepackage{colortbl}
\usepackage{array}

\begin{document}

\title{Generative AI and Large Language Models in Language Preservation:\
Opportunities and Challenges}

\newcommand{\preprintnote}{\textbf{Pre-print version. First published: 20 Jan 2025.}}

\author{Vincent Koc,~\IEEEmembership{Senior Member,~IEEE}%
\thanks{\preprintnote}
\thanks{V. Koc is with Hyperthink Labs, San Francisco, CA 94108 USA (e-mail: vincentkoc@ieee.org).}
\thanks{V. Koc was with the University of Queensland, QLD Australia and the University of New South Wales (UNSW), NSW Australia at the time of this research. He is now affiliated with Comet ML Inc., NY 10003 USA.}
}

\maketitle

\begin{abstract}
The global crisis of language endangerment meets a technological turning point as Generative AI (GenAI) and Large Language Models (LLMs) unlock new frontiers in automating corpus creation, transcription, translation, and tutoring. However, this promise is imperiled by fragmented practices and the critical lack of a methodology to navigate the fraught balance between LLM capabilities and the profound risks of data scarcity, cultural misappropriation, and ethical missteps. This paper introduces a novel analytical framework that systematically evaluates GenAI applications against language-specific needs, embedding community governance and ethical safeguards as foundational pillars. We demonstrate its efficacy through the Te Reo Māori revitalization, where it illuminates successes, such as community-led Automatic Speech Recognition achieving 92\% accuracy, while critically surfacing persistent challenges in data sovereignty and model bias for digital archives and educational tools. Our findings underscore that GenAI can indeed revolutionize language preservation, but only when interventions are rigorously anchored in community-centric data stewardship, continuous evaluation, and transparent risk management. Ultimately, this framework provides an indispensable toolkit for researchers, language communities, and policymakers, aiming to catalyze the ethical and high-impact deployment of LLMs to safeguard the world's linguistic heritage.
\end{abstract}

\begin{IEEEImpStatement}
The accelerating loss of endangered languages poses a critical threat to cultural diversity and heritage worldwide. This paper introduces a practical framework for leveraging Generative AI (GenAI) and Large Language Models (LLMs) to support language preservation efforts. By guiding researchers, language communities, and policymakers in responsibly applying these technologies, our approach enables the creation of digital archives, educational resources, and linguistic tools tailored to community needs. Importantly, the framework emphasizes ethical engagement and cultural sensitivity, helping to ensure that technological interventions benefit language speakers and respect their traditions. Adoption of this methodology can empower communities, inform policy, and foster more inclusive and sustainable language revitalization initiatives.
\end{IEEEImpStatement}


\section{Introduction}
\IEEEPARstart{L}{anguage} is not only a means of communication but also a means of preserving the personality, culture, and traditions of its users. The world's linguistic diversity is immense, yet starkly imbalanced: an estimated 94\% of the global population speaks only 6\% of its languages\cite{NatGeoLDI}. In Turkish, there is a proverb ``Bir lisan bir insan,'' which literally translates to ``one language, one person.'' This means that as you learn a language you embody not only that form of communication but its culture and tradition which is very diverse. However, the progress of globalization, technological change, and social change has led to the extinction of many languages, and many languages are at risk of extinction. UNESCO says that 40 percent of the world's languages are endangered, and Bromham~\emph{et al.}~\cite{Bromham2022} project that, without intervention, language loss could triple within 40 years, with at least one language disappearing per month. This highlights the urgency of preservation efforts. The United Nations, through bodies like UNESCO, emphasizes the right to freedom of expression in a language of one's choice, promoting multilingualism and the development of multilingual digital content and systems \cite{UNESCO2003Rec}. Language endangerment leads to the loss of values and knowledge that cannot be easily replaced\cite{Abiri2024}. Interestingly, while factors like increased road density (facilitating population movement) and higher average years of formal schooling correlate with increased endangerment, direct contact with other languages is not, in itself, a primary driver of language loss\cite{Bromham2022}.

Traditional methods of language preservation, such as field photography, oral history recording, and dictionaries, have been effective in documenting endangered languages. However, these methods often fail to achieve the desired results due to limited resources, limited time, and limited native speaker skills. In recent years, there has been great interest in using artificial intelligence and large-scale language models (LLMs) as tools for language preservation and revitalization. These AI-driven technologies are not limited to textual content but also provide opportunities to bridge the gap between traditional languages and modern technological tools.

Generative AI, including models such as GPT-3, has demonstrated the ability to generate coherent speech in multiple languages. By training these models on existing information, they can help preserve and revitalize endangered languages in various ways. However, challenges remain, particularly regarding data gaps, technical limitations, bias, and ethical issues surrounding ownership and representation\cite{Ajuzieogu, Koc2025Fairness}. This article seeks to explore the opportunities that artificial intelligence offers for language preservation and critically examine the challenges they present.

\section{Background}
In the linguistic context, generative AI systems often rely on deep learning architectures, particularly transformers~\cite{Vaswani2017}, which have revolutionized natural language processing (NLP). Models such as BERT, GPT, and T5 can process large strings of text and capture relationships between words, phrases, and sentences. For a broader overview of recent advances in these technologies, readers are referred to works such as Hagos~\emph{et al.}~\cite{Hagos2024TAI}.

High-adoption language models such as OpenAI's GPT-3 and GPT-4~\cite{Schillaci2024} are built on transformer architectures and are trained extensively on text from the internet, books, and other sources to produce coherent, contextually relevant text. For endangered languages, training these models requires large amounts of high-quality data. While some languages have abundant digital corpora, many endangered languages do not. To address this, AI researchers have begun collecting data from oral traditions, folktales, interviews, and audio recordings. Speech technology is leveraged to transcribe spoken content, which LLMs can then process to create text-based content\cite{Bentalha2024}.

In addition, models such as GPT-3 are often pretrained on large general-purpose corpora and then fine-tuned for specific languages or tasks. This fine-tuning allows the model to adapt to specialized vocabularies, grammar, and cultural contexts, even in endangered languages. The ability to train models on smaller datasets, especially with transfer learning, has proven to be crucial in language preservation where resources are limited. The next section formalises how these capabilities are mapped onto endangered-language needs.

\section{Methodology}\label{sec:methodology}
This paper employs an analytical methodology to comprehensively evaluate the role, potential, and challenges of Generative AI (GenAI) and Large Language Models (LLMs) in the context of endangered language preservation. Our approach is structured around several key investigative pillars, as illustrated in Figure~\ref{fig:taxonomy_opp_chal}:

\begin{figure}[t]
    \centering
    \resizebox{0.48\textwidth}{!}{%
        \begin{tikzpicture}[
          grow=east,
          edge from parent path={(\tikzparentnode.east) -- (\tikzchildnode.west)},
          every node/.style={rectangle, rounded corners, draw, thick, font=\sffamily\small, align=center, anchor=center},
          root/.style={fill={rgb:black,1;white,10}, text width=3.2cm, minimum height=1.2cm},
          category/.style={text width=2.8cm, minimum height=1.1cm},
          opportunity/.style={category, fill=green!10},
          challenge/.style={category, fill=red!10},
          leaf/.style={text width=2.8cm, minimum height=1.1cm, font=\sffamily\footnotesize, draw=black!60},
          level 1/.style={sibling distance=6cm, level distance=4.5cm},
          level 2/.style={sibling distance=1.2cm, level distance=4.5cm}
        ]
        \node[root] {GenAI in \\Language \\ Preservation}
          child { node[challenge] {Challenges}
            child { node[leaf] {Data Scarcity} }
            child { node[leaf] {Technical \\ Limitations} }
            child { node[leaf] {Ethical \\ Considerations} }
            child { node[leaf] {Risk of Cultural \\ Dilution} }
          }
          child { node[opportunity] {Opportunities}
            child { node[leaf] {Language Archiving \\ Digital Preservation} }
            child { node[leaf] {Revitalizing Languages \\ via Education} }
            child { node[leaf] {Enhanced Communication} }
            child { node[leaf] {Supporting \\ Linguistic Research} }
            child { node[leaf] {Speech Recognition \\ Documentation} }
          };
        \end{tikzpicture}
    }
    \caption{Taxonomy of Opportunities and Challenges in Applying Generative AI to Language Preservation.}\label{fig:taxonomy_opp_chal}
\end{figure}

\subsection{Literature Review and Synthesis}\label{sec:methodology_lit_review} We conduct a review of current literature on GenAI/LLMs (as detailed in the Background section) and existing language preservation efforts. This establishes the technological state-of-the-art and current preservation paradigms.

\subsection{Framework for Opportunity and Challenge Identification}\label{sec:methodology_framework_opp_chal} We develop and apply a systematic framework to identify, categorize, and analyze the opportunities presented by GenAI for language preservation. This includes areas such as documentation, education, revitalization, and communication. Concurrently, the framework addresses the spectrum of challenges, including data scarcity, technical limitations, computational resource demands, and the risk of linguistic or cultural inaccuracies. The core components and process of this analytical framework are defined in Figure~\ref{fig:analytical_framework_definition}.

    \begin{figure}[htb]
    \centering
    \fbox{
    \begin{minipage}{0.9\columnwidth}
    \centering
    Analytical Framework Overview
    \vspace{0.5em}
    \hrule
    \vspace{0.5em}
    \begin{flushleft}
    Inputs:
    \begin{itemize}
        \item Target Endangered Language: \(L_i \in \mathcal{L}\)
        \item Set of GenAI Technological Capabilities: \(\mathcal{T}_{AI} = \{t_1, t_2, \ldots t_n\}\)
        \hspace{1em}(e.g., \(t_1\) = text generation, \(t_2\) = speech-to-text transcription)
    \end{itemize}
    Process:
    \begin{enumerate}
        \item[1.] Systematically analyze the application of each technological capability in \(\mathcal{T}_{AI}\) to the specific context of language \(L_i\).
        \item[2.] Identify and categorize potential opportunities.
        \item[3.] Identify and categorize potential challenges.
    \end{enumerate}
    Functional Mapping Summary:
    \begin{align*}
    (L_i, \mathcal{T}_{AI}) &\xrightarrow{\text{Anal.}} (O(L_i, \mathcal{T}_{AI}), C(L_i, \mathcal{T}_{AI})) \\
    & \hspace{2em} \xrightarrow{\text{Synth.}} S(L_i)
    \end{align*}
    Outputs:
    \begin{itemize}
        \item Set of Identified Opportunities: \(O(L_i, \mathcal{T}_{AI})\)
        \item Set of Identified Challenges: \(C(L_i, \mathcal{T}_{AI})\)
        \item Set of Recommended Strategies: \(S(L_i)\), derived from synthesizing \(O(L_i, \mathcal{T}_{AI})\) and \(C(L_i, \mathcal{T}_{AI})\).
    \end{itemize}
    \end{flushleft}
    \end{minipage}
    }
    \caption{The proposed analytical framework detailing inputs, core processes (numbered 1 to 3 to visually echo the text), a functional mapping summary, and outputs for assessing GenAI applications in language preservation.}\label{fig:analytical_framework_definition}
    \end{figure}

\subsection{Ethical and Cultural Impact Assessment}\label{sec:methodology_ethical_cultural} A critical component of our methodology is the in-depth examination of the ethical and cultural implications. This involves analyzing issues of data sovereignty, community engagement, intellectual property, the potential for bias amplification, and the importance of culturally sensitive AI development and deployment.

\subsection{Case Study Analysis}\label{sec:methodology_case_study} We analyze illustrative case studies where AI technologies have been applied to language preservation efforts. These cases are not exhaustive but serve to ground the theoretical discussion in practical examples, highlighting successes, limitations, and lessons learned.

\subsection{Synthesis and Guideline Formulation}\label{sec:methodology_synthesis_guidelines} Finally, the findings from the above steps are synthesized to provide a holistic understanding. This synthesis forms the basis for proposing considerations and guiding principles for the responsible and effective use of GenAI, maximizing benefits for language revitalization while mitigating risks of cultural misrepresentation, fostering equitable technological engagement for global linguistic diversity preservation.

This methodological approach allows for a multifaceted exploration of the topic, moving beyond a purely technical assessment to include critical socio-cultural and ethical dimensions essential for sustainable and respectful language preservation. Section~\ref{sec:framework_application} walks through the application of this framework to the Te Reo Māori case study.

\section{Framework Application: A Worked Example with Te Reo Māori}\label{sec:framework_application}
We now walk the framework on Te Reo Māori, the leading revival effort in Aotearoa New Zealand. Te Reo Māori has experienced a significant revival, with AI technology playing an increasingly supportive role. The table below (Table~\ref{tab:framework_application_maori}) provides a condensed overview of how the framework components map to the specifics of the Te Reo Māori case. A more detailed narrative of this application is provided in Appendix~\ref{app:maori_detail}.

\begin{table*}[!htbp]
    \centering
    \caption{Summary of Analytical Framework Application to Te Reo Māori Revitalization Efforts}
    \label{tab:framework_application_maori}
    \resizebox{\textwidth}{!}{%
    \begin{tabular}{@{} >{\raggedleft\arraybackslash}p{0.25\textwidth}lp{0.55\textwidth}@{}}
    \toprule
    \multicolumn{2}{@{}l}{\textbf{Framework Component (from Fig.~\ref{fig:analytical_framework_definition})}} & \textbf{Application to Te Reo Māori (Summary)} \\
    \midrule
    \multicolumn{2}{@{}l}{\textbf{Inputs}} & \\
    & Target Language (\(L_i\)) & Te Reo Māori \\
    & GenAI Capabilities (\(\mathcal{T}_{AI}\)) & Automatic Speech Recognition (ASR), Natural Language Generation (NLG), Data Management/Augmentation. Examples include Te Hiku Media's ASR models \cite{Time2024}. \\
    \addlinespace
    \multicolumn{2}{@{}l}{\textbf{Process}} & \\
    & 1. Analyze Tech Application to \(L_i\) & Collaborative needs assessment with the Māori community, evaluation of AI model suitability for Te Reo Māori's linguistic features, culturally sensitive data collection and ethical planning. \\
    & 2. Identify Opportunities & Enhanced language archiving (digital transcription of oral histories), development of interactive learning tools (AI-powered apps, chatbots), creation of new Māori language content, and support for linguistic research. \\
    & 3. Identify Challenges & Data scarcity and quality for a less-resourced language, technical complexity in capturing linguistic nuances, ensuring cultural authenticity and avoiding misrepresentation, upholding community ownership and data sovereignty, and addressing resource demands (computational, financial, personnel). \\
    \addlinespace
    \multicolumn{2}{@{}l}{\textbf{Outputs}} & \\
    & Opportunities (\(O(L_i, \mathcal{T}_{AI})\)) & Summarized as: Creation of accessible digital archives, engaging educational resources, increased language visibility, and advanced research capabilities. \\
    & Challenges (\(C(L_i, \mathcal{T}_{AI})\))    & Summarized as: Need to overcome data limitations, technical hurdles, cultural/ethical risks, and resource constraints. \\
    & Strategies (\(S(L_i)\))               & Community-led development and governance, capacity building within the Māori community, adoption of ethical AI guidelines prioritizing cultural protocols, iterative development with human oversight, and focus on low-resource AI techniques. \\
    \bottomrule
    \end{tabular}%
    }
\end{table*}

This summarized application demonstrates the framework's utility in structuring the analysis of GenAI projects for language preservation, ensuring a comprehensive consideration of inputs, processes, and the resulting opportunities, challenges, and strategies. For a detailed narrative of each step, please refer to Appendix~\ref{app:maori_detail}.

\section{Applying Generative AI in Language Preservation: Analysis and Insights}\label{sec:analysis_insights}
While Section~\ref{sec:framework_application} provided a detailed worked example with Te Reo Māori, the framework and insights are applicable more broadly. The challenges and opportunities identified resonate across various language preservation contexts. For example, the Sami languages in Northern Europe also face endangerment. GenAI opportunities include developing educational tools accessible across dispersed Sami communities and aiding in the documentation of diverse dialects. Challenges involve data scarcity for specific dialects, ensuring representation of all Sami varieties, and navigating data governance across multiple nation-states where Sami people reside. Ethical considerations include respecting Sami cultural protocols around knowledge sharing and ensuring that AI tools support, rather than supplant, traditional teaching methods and intergenerational transmission.

\subsection{Opportunities in AI-driven Language Preservation}
Generative AI introduces numerous opportunities to complement traditional language preservation methods. One of the primary advantages is the ability to produce large volumes of language materials (written or spoken) almost instantly\cite{Chen2024}. AI-driven systems can generate text, audio, and even video that can be used in educational efforts, cultural revitalization, and digital
archives.

\textbf{Language Archiving and Digital Preservation:} AI can assist in archiving endangered languages by creating digital repositories of written and spoken texts. For spoken languages, AI models can transcribe audio recordings, preserving the language in written form. This transcription process can be automated, reducing the workload on linguists and field researchers. For instance, AI-generated text can help linguists develop dictionaries, grammar guides, and phonetic notation---all critical tools for language preservation\cite{Feretzakis2024}.

\textbf{Revitalizing Languages through Education:} AI-powered language learning platforms have the potential to revitalize endangered languages by equipping a new generation of speakers. Generative AI can be used to create language-learning applications, such as chatbots and virtual tutors, that simulate real-life conversations in the target language. These tools enable learners to practice speaking and understanding the language in an immersive, interactive environment. Additionally, AI can be employed to generate customized learning materials, such as reading passages, exercises, and quizzes, to address the individual needs of learners.

\textbf{Enhanced Communication:} For minority language speakers, AI systems can enable communication with majority-language speakers. AI-powered translation systems can bridge linguistic gaps, allowing speakers of endangered languages to communicate with wider audiences. Such translations are particularly valuable for communities dispersed globally or with limited access to language education\cite{Hadi2024}.

\textbf{Supporting Linguistic Research:} AI can support language research by identifying patterns and structures in languages that might not be immediately evident to human researchers. This is especially important for lesser-known languages, where data may be sparse or fragmented. By processing large datasets, AI can reveal syntactic, morphological, and phonetic features that can lead to a deeper understanding of a language.

\textbf{Speech Recognition and Documentation:} In unwritten languages, generative AI can aid in speech recognition and processing. AI systems can be trained to identify spoken patterns and convert them into text. This presents new opportunities to document and preserve the lexicon of endangered languages, which is critical for safeguarding unique phonetic and pronunciation features. For instance, Te Hiku Media developed an automatic speech recognition (ASR) model for the Māori language, Te Reo Māori, achieving 92\% accuracy in transcription (corresponding to a low Word Error Rate, e.g., 8\% WER, compared to baselines which might be 20\% WER or higher), outperforming similar attempts by major international tech companies\cite{Time2024}. Similarly, researchers at Dartmouth have successfully built automatic speech-recognition models for Cook Islands Māori that effectively identify speech patterns from audio recordings and transcribe them into text\cite{Dartmouth2025}. These examples demonstrate the potential of AI-driven speech recognition in preserving endangered languages, particularly when developed in collaboration with language communities and leveraging existing archival materials.

\subsection{Challenges in AI-driven Language Preservation}
Despite its vast potential, generative AI also poses several challenges that must be addressed to ensure effective and ethical language preservation.

\textbf{Data Scarcity:} One of the most significant hurdles is the lack of sufficient data in many endangered languages. While major languages such as English, Spanish, and Mandarin boast enormous digital footprints, most endangered languages lack comprehensive written records. This deficit constrains the training of robust AI models. Even when data do exist, they may be fragmented, inconsistent, or incomplete, complicating the model training process\cite{Khan2024}. Addressing such data limitations is a crucial research area, with techniques like text data augmentation showing promise in related NLP fields\cite{Abonizio2022TAI}, and methods for adapting multilingual LLMs to low-resource languages using knowledge graphs and adapters offering new avenues for improvement\cite{gurgurov-etal-2024-adapting}.

\textbf{Technical Limitations:} Training large-scale language models requires significant computational resources, including specialized hardware and large amounts of electricity. Recent research has shown that AI workloads demand substantial resources, particularly specialized hardware accelerators like TPUs and GPUs\cite{Sanmartin2024}. The development of advanced AI models has demonstrated that while training costs can be optimized, the shift towards more sophisticated reasoning models significantly increases computational demands\cite{FT2024}. Access to the right infrastructure can be an obstacle, especially for endangered languages spoken in low-resource regions. Recognizing this challenge, initiatives like the National Artificial Intelligence Research Resource (NAIRR) pilot program have been launched to democratize access to computational power and AI research tools\cite{Time2024NAIRR}. Additionally, AI models often struggle with complex grammar, non-standard spellings, or extensive lexical borrowing from dominant regional languages, leading to less-accurate outputs. Research has shown that multilingual AI systems face significant challenges in balancing performance across diverse languages, particularly those with complex grammatical structures and regional variations\cite{app15073882, Koc202XNonLatinLLMEvalComet}. These limitations are further compounded by the tendency of AI models to generate plausible but incorrect content when dealing with unfamiliar linguistic patterns\cite{PNAS2023}.

\textbf{Risk of Cultural Dilution:} While AI is a powerful tool for language revitalization, it can inadvertently contribute to the dilution of language and cultural identity. Automating language learning or generation may diminish the richness and authenticity that comes from human speakers who carry cultural histories in their speech. Recent studies have demonstrated that AI-generated content often lacks the nuanced cultural context and authenticity present in human-created materials\cite{Wani2024TAI}. AI models, optimized for efficiency and coherence, could inadvertently overlook deep-rooted cultural contexts. Research has shown that AI's understanding of cultural context is inherently limited, as it lacks the lived experiences and emotional intelligence needed to fully grasp cultural subtleties\cite{CampoRuiz2025, heritage7110287}. The increasing sophistication of AI in generating human-like text also brings challenges in distinguishing AI-generated content from authentic community-created materials, raising concerns about the preservation of cultural authenticity in language documentation\cite{PNAS2023}. This challenge is particularly acute for endangered languages, where the loss of even subtle cultural nuances can significantly impact the preservation of linguistic heritage.

\subsection{Ethical and Cultural Dimensions}\label{sec:ethical_cultural_dimensions}
The application of AI in language preservation necessitates careful navigation of ethical and cultural landscapes. Key considerations from previous sections will be integrated and expanded here. For instance, using AI for language preservation raises important ethical questions, particularly regarding ownership and cultural sensitivity of linguistic data. Communities whose languages are endangered may have little control over how their languages are represented or monetized. As emphasized by Peter-Lucas Jones, CEO of Te Hiku Media, indigenous communities must maintain control and governance of their digital data to avoid digital disenfranchisement \cite{Time2024}. This proactive, community-centered approach, as exemplified in the Te Reo Māori case (see Section~\ref{sec:framework_application} and Appendix~\ref{app:maori_detail}), is crucial for ensuring that technological interventions respect data sovereignty and mitigate risks such as digital colonialism or the perpetuation of biases present in training data. For instance, open-sourcing language materials without community consent can be counterproductive, as it may lead to unintended consequences in the broader geopolitical context of AI development and national security\cite{BusinessInsider2025}. Moreover, AI-generated texts might not capture the full cultural or linguistic nuance, risking distortion or misrepresentation\cite{Leslie2024}.

\begin{figure}[t]
    \centering
    \includegraphics[width=\columnwidth]{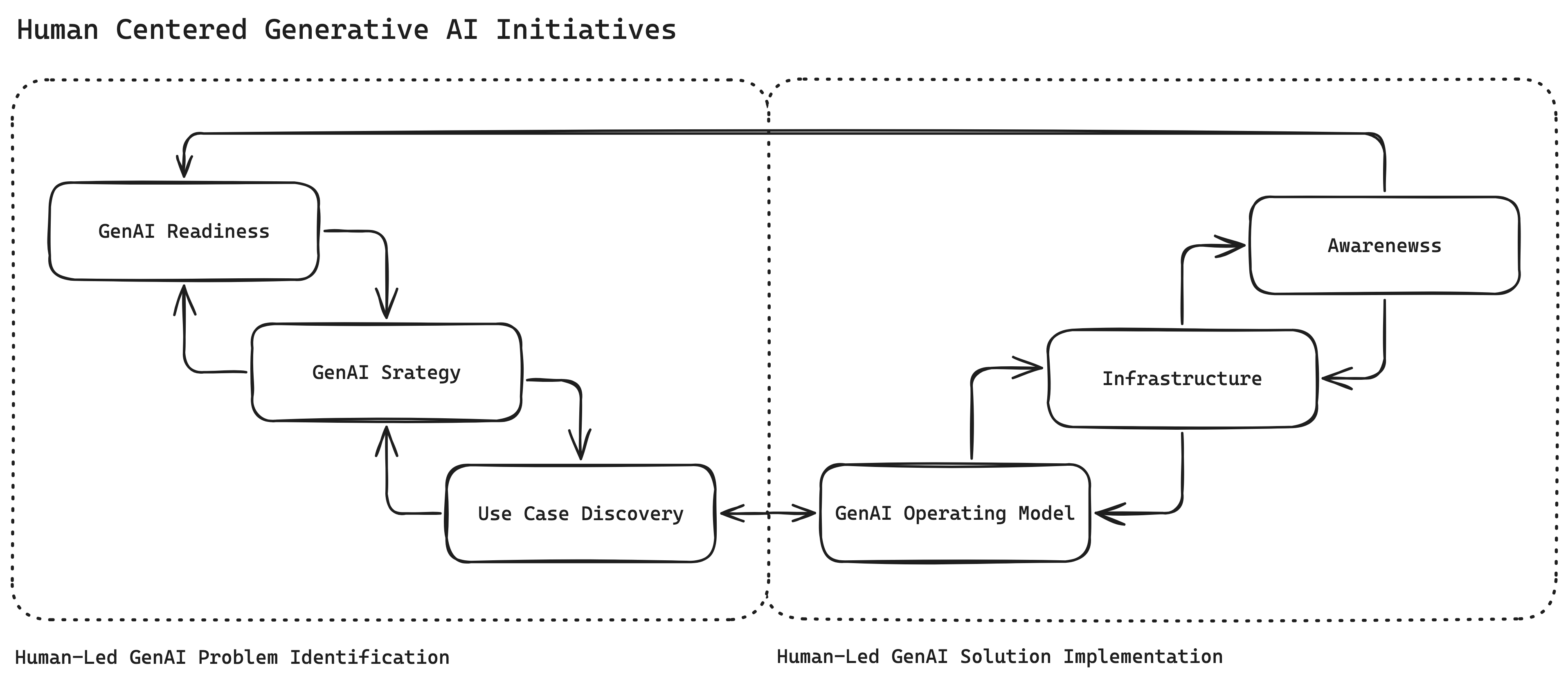}
    \caption{A human-centered framework for GenAI initiatives, illustrating the dual phases of problem identification and solution implementation, with iterative feedback between readiness, strategy, use case discovery, operating model, infrastructure, and awareness. Adapted and redesigned based on PwC~\cite{pwc2025}.}\label{fig:initiative_tikz}
\end{figure}

As illustrated in Figure~\ref{fig:initiative_tikz}, a human-centered approach to GenAI initiatives involves both problem identification and solution implementation phases, ensuring that readiness, strategy, and use case discovery are tightly integrated with operational, infrastructural, and awareness-building activities.

Informed consent is paramount: the collection and use of linguistic data should be done with community permission and oversight, ensuring respect for indigenous knowledge and preventing misuse. Likewise, cultural sensitivity is crucial. AI models should be trained to respect the cultural dimensions embedded within a language; otherwise, deeper meanings might be lost or misconstrued\cite{Mizrahi2024}. The UNESCO Recommendation concerning the Promotion and Use of Multilingualism and Universal Access to Cyberspace further calls for collaboration among international organizations, governments, civil society, academia, and the private sector to develop multilingual content and systems, facilitate access, and ensure an equitable balance between the interests of rights-holders and the public interest\cite{UNESCO2003Rec}.

\section{Proposed ImpactScore Rubric for Intervention Assessment}\label{sec:impact_score_rubric}
To aid in the strategic selection and prioritization of GenAI-driven language preservation projects, we propose a conceptual multi-criteria assessment rubric, the \texttt{ImpactScore}. This rubric, inspired by frameworks such as the AI for IMPACTS framework\cite{PMC2024}, aims to provide a structured approach for evaluating the potential impact and feasibility of different AI interventions (\(I_k\)) for a specific target language (\(L_i\)). The assessment can be conceptualized as:

    \begin{equation*}
    \begin{split}
    \text{ImpactScore}(I_k, L_i) = f(& \text{OpportunityFit}, \\
                                    & \text{DataAvailability}, \\
                                    & \text{CommunitySupport}, \\
                                    & \text{EthicalRiskLevel}, \\
                                    & \text{ResourceNeeds})
    \end{split}
    \end{equation*}
where \(f(\ldots)\) represents a function that synthesizes these weighted critical factors. Key components include:

\begin{itemize}
    \item \textbf{OpportunityFit:} How well the proposed intervention aligns with the identified needs and strategic goals for the preservation/revitalization of language \(L_i\).
    \item \textbf{DataAvailability:} The current state and accessibility of linguistic data (corpora, lexical resources, audio-visual materials) necessary to train and implement the AI model for \(I_k\).
    \item \textbf{CommunitySupport:} The level of engagement, endorsement, and active participation from the language community in the proposed intervention.
    \item \textbf{EthicalRiskLevel:} An assessment of potential ethical risks (e.g., bias, misrepresentation, data sovereignty issues) and the robustness of mitigation strategies.
    \item \textbf{ResourceNeeds:} The technical infrastructure, financial investment, and human expertise required for the successful development, deployment, and sustainability of \(I_k\).
\end{itemize}
This framework provides a structured approach to decision-making, ensuring that interventions are both technically feasible and culturally appropriate. An illustrative application of this rubric to a hypothetical Te Reo Māori intervention is detailed in Appendix~\ref{app:impact_score_maori}.

\section{Conclusion and Future Work}\label{sec:conclusion}
Generative AI (GenAI) and LLMs present a paradigm with transformative potential for the preservation and revitalization of endangered languages. This paper has introduced an analytical methodology for systematically evaluating and applying these advanced AI technologies, navigating their capabilities against significant challenges such as data scarcity, technical hurdles, and crucial ethical and cultural considerations. Our analysis, structured through this methodology and demonstrated with the Te Reo Māori case, highlights that while GenAI offers powerful tools, their deployment necessitates a carefully considered, community-centric approach.

The core contribution of this work is the proposed methodological framework, which encourages a balanced perspective, harnessing the technological power of AI while foregrounding ethical integrity and cultural sensitivity. Effective and respectful language preservation using AI hinges on robust collaboration between AI developers, linguists, heritage-bearers, and the language communities themselves. The proposed \texttt{ImpactScore} rubric (Section~\ref{sec:impact_score_rubric}) further offers a tool for strategic project evaluation.

Looking ahead, several specific research directions are critical to advancing this field responsibly:
\begin{itemize}
    \item Low-Resource Learning Techniques: Developing and refining AI models that can perform effectively with the limited datasets typical of endangered languages remains a primary challenge. Future work should focus on transfer learning, few-shot learning, and data augmentation techniques (e.g., extending approaches like those in \cite{Abonizio2022TAI}) specifically adapted to the unique linguistic structures and cultural contexts of these languages, including promising approaches like knowledge graph integration via adapters \cite{gurgurov-etal-2024-adapting}.
    \item Interpretable and Explainable AI (XAI): Enhancing the interpretability of LLMs used for language preservation is crucial for building trust and facilitating collaboration with language communities. Research into XAI methods can help demystify how models process and generate language, enabling linguists and community members to better understand, guide, and validate AI-generated content.
    \item Community-Centric AI Frameworks: Future research should prioritize the development of co-design frameworks where language communities are integral partners throughout the AI development lifecycle, from data collection and model training to application design and deployment. This includes creating tools and platforms that empower communities to manage their own linguistic data and AI initiatives. This aligns with calls for urgent investment in community-based programs, alongside language documentation and bilingual education, to avert substantial language loss\cite{Bromham2022}.
    \item Cultural Resonance and Authenticity Metrics: There is a need for novel evaluation metrics that go beyond standard NLP accuracy to assess the cultural resonance, appropriateness, and authenticity of AI-generated language content. This may involve interdisciplinary approaches combining computational linguistics with anthropology, sociology, and indigenous studies.
    \item Ethical Guidelines and Governance Models: Continued research is needed to establish comprehensive ethical guidelines and governance models for the use of AI in language preservation. These should address data sovereignty, intellectual property rights, benefit-sharing, and the prevention of digital colonialism, ensuring that AI serves the true interests of the language communities.
\end{itemize}

By pursuing these future research avenues, the global community can better ensure that the advancements in GenAI and LLMs contribute positively and equitably to the vital mission of safeguarding the world's linguistic diversity for generations to come.

\appendix

\section{Detailed Application of the Analytical Framework to Te Reo Māori}\label{app:maori_detail}
This appendix provides a more detailed narrative of the application of the analytical framework (presented in Figure~\ref{fig:analytical_framework_definition}) to the revitalization efforts for Te Reo Māori, as summarized in Table~\ref{tab:framework_application_maori} in Section~\ref{sec:framework_application}.

\subsection{Context: Te Reo Māori Revitalization}
Te Reo Māori (the Māori language) in New Zealand has experienced a significant revival, with AI technology playing an increasingly supportive role. Community initiatives and researchers have developed AI systems, contributing to new educational resources and the preservation of cultural heritage. Machine learning models, for example, have been employed to transcribe audio recordings of Māori oral traditions, aiding in the creation of digital archives\cite{Time2024}.

\subsection{Inputs: Language and Technological Capabilities}
In applying the framework, the target endangered language (\(L_i\)) is Te Reo Māori. The primary set of GenAI technological capabilities (\(\mathcal{T}_{AI}\)) utilized in these initiatives includes:
\begin{itemize}
    \item Automatic Speech Recognition (ASR): For transcribing spoken Māori from audio and video archives, and for interactive speaking practice tools. This is exemplified by Te Hiku Media's work, which achieved high accuracy by training models on community data\cite{Time2024}.
    \item Natural Language Generation (NLG): For creating educational content, generating language examples, and potentially for chatbot-based learning applications. This can help produce new teaching materials and interactive exercises.
    \item Data Management and Augmentation Techniques: To handle and enhance often scarce or fragmented linguistic data specific to Māori. This includes methods for cleaning, organizing, and potentially augmenting datasets to improve model performance.
\end{itemize}

\subsection{Process: Framework-Guided Analysis}
The application of these GenAI capabilities to Te Reo Māori, guided by the framework, necessitates a systematic analysis. This involves several key steps:
\begin{enumerate}
    \item Assessing the specific needs and goals of the Māori language community in collaboration with its members. This initial step is crucial for ensuring that technological interventions align with community aspirations. The community-centric approach of initiatives like Te Hiku Media, which ensured ASR models were trained on community data and respected indigenous data sovereignty, serves as a prime example\cite{Time2024}.
    \item Evaluating the suitability and adaptability of existing GenAI models for Te Reo Māori, considering its unique linguistic features (e.g., phonology, morphology, syntax). This may involve pilot projects or feasibility studies.
    \item Planning for data collection, curation, and annotation in a culturally sensitive manner. This includes addressing potential data gaps, establishing protocols for data governance and ownership, and ensuring ethical use throughout the data lifecycle.
\end{enumerate}

\subsection{Outputs and Insights: Opportunities, Challenges, and Strategies}
Applying the framework systematically helps to identify and categorize the following outputs and insights related to the Te Reo Māori case:

\subsubsection{Identified Opportunities (\(O(L_i, \mathcal{T}_{AI})\))}
The use of GenAI in Te Reo Māori revitalization presents manifold and significant opportunities:
\begin{itemize}
    \item Enhanced Language Archiving and Accessibility: Digital transcription of oral histories, traditional narratives, and everyday speech makes invaluable linguistic and cultural knowledge more accessible for current and future generations and more durable than physical media.
    \item Development of Interactive and Personalized Learning Tools: AI-powered applications, chatbots, and educational games can make learning Te Reo Māori more engaging, adaptive to individual learner needs, and available to a wider audience.
    \item Creation and Dissemination of New Māori Language Content: Generation of texts, scripts, subtitles, and educational materials for various platforms (online, media, publishing) increases the language's visibility, prestige, and utility in modern digital contexts.
    \item Support for Linguistic Research: AI tools can assist linguists in analyzing complex linguistic patterns, tracking language change and variation, supporting lexicography, and developing grammatical resources.
\end{itemize}

\subsubsection{Identified Challenges (\(C(L_i, \mathcal{T}_{AI})\))}
Despite the potential, significant challenges must be acknowledged and navigated:
\begin{itemize}
    \item Data Scarcity and Quality: Like many endangered languages, Te Reo Māori faced an initial lack of large, high-quality, and diverse digital corpora necessary for training robust AI models. Existing data may be fragmented or inconsistently annotated.
    \item Technical Complexity and Accuracy: Accurately capturing the nuances of Māori phonetics, grammar, dialects, and cultural context with general-purpose AI models requires significant adaptation and specialized expertise.
    \item Cultural Authenticity and Risk of Misrepresentation: There is a risk that AI-generated content may not be culturally appropriate, may oversimplify linguistic richness, or could inadvertently perpetuate biases or inaccuracies if not carefully managed.
    \item Community Ownership, Data Sovereignty, and Ethical Concerns: Critical questions arise regarding who owns the language data, how it is collected, stored, and used, and ensuring that benefits from AI initiatives return to and empower the Māori community. The Te Hiku Media example underscores the importance of community control over their data\cite{Time2024}.
    \item Resource Demands and Sustainability: Access to computational resources, sustainable funding, and skilled personnel (both technical and linguistic) are necessary for the long-term development, maintenance, and evolution of AI tools for language revitalization.
\end{itemize}

\subsubsection{Recommended Strategies (\(S(L_i)\))}
The framework helps in deriving and highlighting effective strategies that have been or should be employed:
\begin{itemize}
    \item Community-Led Development and Governance: Prioritizing deep collaboration with, and often leadership from, the Māori community at all stages, from project inception and design to development, deployment, and evaluation.
    \item Capacity Building within the Community: Investing in training and empowering community members to develop, manage, and utilize AI tools for their language, fostering self-determination and sustainability.
    \item Adoption of Ethical AI Practices and Indigenous Data Sovereignty Principles: Establishing clear guidelines for data governance, intellectual property rights, benefit-sharing, and culturally sensitive AI development that prioritize community protocols and values.
    \item Iterative Development with Human-in-the-Loop Oversight: Combining the capabilities of AI-generated content with rigorous review, refinement, and validation by fluent speakers, elders, and cultural experts to ensure quality, accuracy, and authenticity.
    \item Focus on Low-Resource Techniques and Transfer Learning: Exploring and adapting AI methodologies that are suitable for languages with limited digital data, leveraging knowledge from higher-resource languages where appropriate but with careful adaptation.
\end{itemize}
This detailed application to Te Reo Māori illustrates how the systematic use of the analytical framework can help stakeholders identify key considerations, anticipate challenges, and formulate robust strategies for leveraging GenAI in language preservation. It underscores that technological interventions must be effective, ethical, and culturally appropriate to maximize benefits and mitigate risks.

\subsection{Illustrative ImpactScore Assessment for a Te Reo Māori Intervention}\label{app:impact_score_maori}
Building upon the framework application, we can illustrate the potential use of the \texttt{ImpactScore} concept introduced in Section~\ref{sec:impact_score_rubric}. The \texttt{ImpactScore} is intended as a multi-criteria assessment to evaluate the potential impact and feasibility of different AI interventions for a specific language, \(L_i\).

For instance, a hypothetical assignment might yield: \texttt{ImpactScore} = 0.78 (on a 0-1 scale, with illustrative weights: OpportunityFit=0.3, DataAvailability=0.2, CommunitySupport=0.3, EthicalRiskLevel=0.1, ResourceNeeds=0.1).

Let us consider a hypothetical GenAI intervention (\(I_k\)) for Te Reo Māori (\(L_i\)): ``Developing and Deploying Community-Centric ASR and NLG Tools for Enhanced Archiving and Interactive Education.''

The assessment for \(\text{ImpactScore}(I_k, \text{Te Reo Māori}) = f(\ldots)\) would involve evaluating the following factors, with hypothetical qualitative assessments based on the Te Reo Māori case study:

\begin{itemize}
    \item \textbf{OpportunityFit:} \textit{High}. The intervention directly addresses critical needs in language documentation (archiving oral traditions) and revitalization (creating engaging educational tools), aligning well with community goals for preserving and promoting Te Reo Māori.

    \item \textbf{DataAvailability:} \textit{Medium (Improving)}. While initially a challenge (low-resource), concerted efforts by community groups like Te Hiku Media \cite{Time2024} have significantly improved the availability and quality of relevant linguistic data, particularly for ASR. NLG data might still require further development.

    \item \textbf{CommunitySupport:} \textit{High}. Initiatives like Te Hiku Media demonstrate strong community engagement, leadership, and support for leveraging AI in a way that respects Māori data sovereignty and cultural values. This is a crucial positive factor.

    \item \textbf{EthicalRiskLevel:} \textit{Medium (Manageable with Mitigation)}. Risks related to cultural misrepresentation, data misuse, and intellectual property are present. However, the strong community-led approach and emphasis on indigenous data sovereignty (as seen in existing Māori initiatives) provide robust mitigation strategies, making the net risk manageable.

    \item \textbf{ResourceNeeds:} \textit{Medium to High}. Developing and sustaining high-quality, culturally attuned AI tools requires significant and ongoing investment in computational resources, skilled personnel (both technical and linguistic), and funding. While initial successes are evident, long-term sustainability of resources is a key consideration.
\end{itemize}

The function \(f(\ldots)\) would then synthesize these weighted factors. In this illustrative scenario for Te Reo Māori, the high \textit{OpportunityFit} and \textit{CommunitySupport}, coupled with improving \textit{DataAvailability} and manageable \textit{EthicalRiskLevel} (due to strong community governance), would likely lead to a favorable overall \texttt{ImpactScore}. This would indicate that such an intervention is highly promising, provided that the \textit{ResourceNeeds} can be adequately addressed through sustainable funding and capacity-building efforts.

This hypothetical assessment demonstrates how the \texttt{ImpactScore} framework could guide strategic decision-making and prioritization of GenAI projects in language preservation contexts by providing a structured way to consider diverse critical factors.

\section{Key Benefits Table}\label{app:benefits_table}

\begin{table}[!htbp]
    \centering
    \caption{Key Benefits of Generative AI for Endangered Language Documentation, Learning, and Standardization.}
    \label{tab:benefits_appendix_content_actual}
    \begin{tabular}{@{}p{0.28\columnwidth} p{0.62\columnwidth}@{}}
    \toprule
    \textbf{Benefit} & \textbf{Description} \\
    \midrule
    Real-time Generation & Instantly generates content for documentation and teaching. \\
    Oral Language Documentation & Converts speech to text, preserving phonetics and structure. \\
    Language Learning & Simulates conversations for immersive language practice. \\
    Data Standardization & Aids in creating standardized orthographies and phonetic systems. \\
    \bottomrule
    \end{tabular}
\end{table}

\bibliographystyle{IEEEtran}
\bibliography{main}

\begin{IEEEbiography}[{\includegraphics[width=1in,height=1.25in,clip,keepaspectratio]{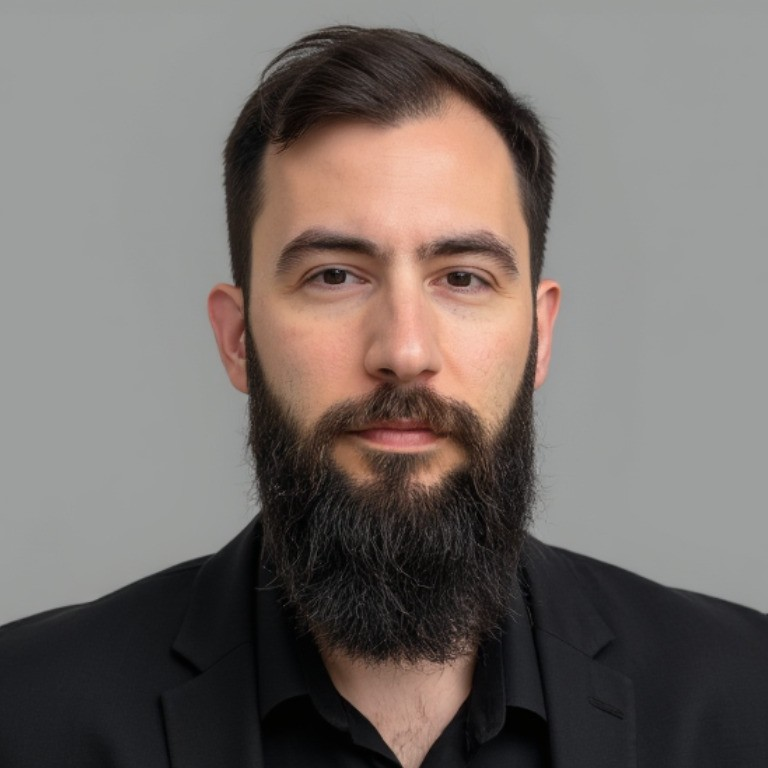}}]{Vincent Koc}{\space}(Senior Member, IEEE) has worked in both industry and academia for over two decades, specializing in artificial intelligence, language technologies, and cross-cultural research. He is the founder of Hyperthink Labs, a profit-for-cause research laboratory that collaborates with leading AI companies worldwide on projects at the intersection of technology and social good.

Vincent's career spans roles in research, engineering, and leadership across multiple continents, including appointments at the University of Queensland, the University of New South Wales, and research positions in industry such as Microsoft. He is a polyglot and multi-national, bringing personal and lived experience to his work on language preservation, computational linguistics, and the ethical deployment of AI. His research is driven by a passion for the intersections of culture, language, and artificial intelligence, and he is committed to advancing responsible and inclusive technology for global communities.

Mr.\ Koc is a Senior Member of the IEEE, a Member of the British Computer Society (BCS), a Member of the Australian Computer Society (ACS), a Member of the National Writers Union (NWU), a Fellow of the Institute of Managers and Leaders (FIML, AUNZ), a Justice of the Peace (NSW, Australia), and a Member of the Forbes Technology Council.
\end{IEEEbiography}

\end{document}